\documentclass[10pt,twocolumn,letterpaper]{article}

\usepackage{cvpr}
\usepackage{times}
\usepackage{epsfig}
\usepackage{graphicx}
\usepackage{amsmath}
\usepackage{amssymb}

\usepackage[utf8]{inputenc} 
\usepackage[ruled,vlined]{algorithm2e}
\usepackage{url}            

\usepackage{graphicx,dblfloatfix}
\usepackage{multirow}

\usepackage{subfig}

\usepackage{amsmath,amssymb} 

\usepackage[pagebackref=true,breaklinks=true,letterpaper=true,colorlinks,bookmarks=false]{hyperref}

\usepackage{fancyhdr,graphicx,amsmath,amssymb}
\usepackage[ruled,vlined]{algorithm2e}
\include{pythonlisting}

\usepackage{color}

\usepackage[breaklinks=true,bookmarks=false]{hyperref}

\cvprfinalcopy 

\def\cvprPaperID{****} 

\ifcvprfinal\fi
\begin{document}

\title{On Learning Density Aware Embeddings}

\author{Soumyadeep Ghosh, Richa Singh, Mayank Vatsa\\
IIIT-Delhi, India\\
{\tt\small \{soumyadeepg, rsingh, mayank\}@iiitd.ac.in}}

\maketitle
\thispagestyle{empty}

\begin{abstract}
    Deep metric learning algorithms have been utilized to learn discriminative and generalizable models which are effective for classifying unseen classes. In this paper, a novel noise tolerant deep metric learning algorithm is proposed. The proposed method, termed as Density Aware Metric Learning, enforces the model to learn embeddings that are pulled towards the most dense region of the clusters for each class. It is achieved by iteratively shifting the estimate of the center towards the dense region of the cluster thereby leading to faster convergence and higher generalizability. In addition to this, the approach is robust to noisy samples in the training data, often present as outliers. Detailed experiments and analysis on two challenging cross-modal face recognition databases and two popular object recognition databases exhibit the efficacy of the proposed approach. It has superior convergence, requires lesser training time, and yields better accuracies than several popular deep metric learning methods.
\end{abstract}

\section{Introduction}
\begin{figure}[h!]
     	\begin{center}
     	\includegraphics[width=.88\linewidth]{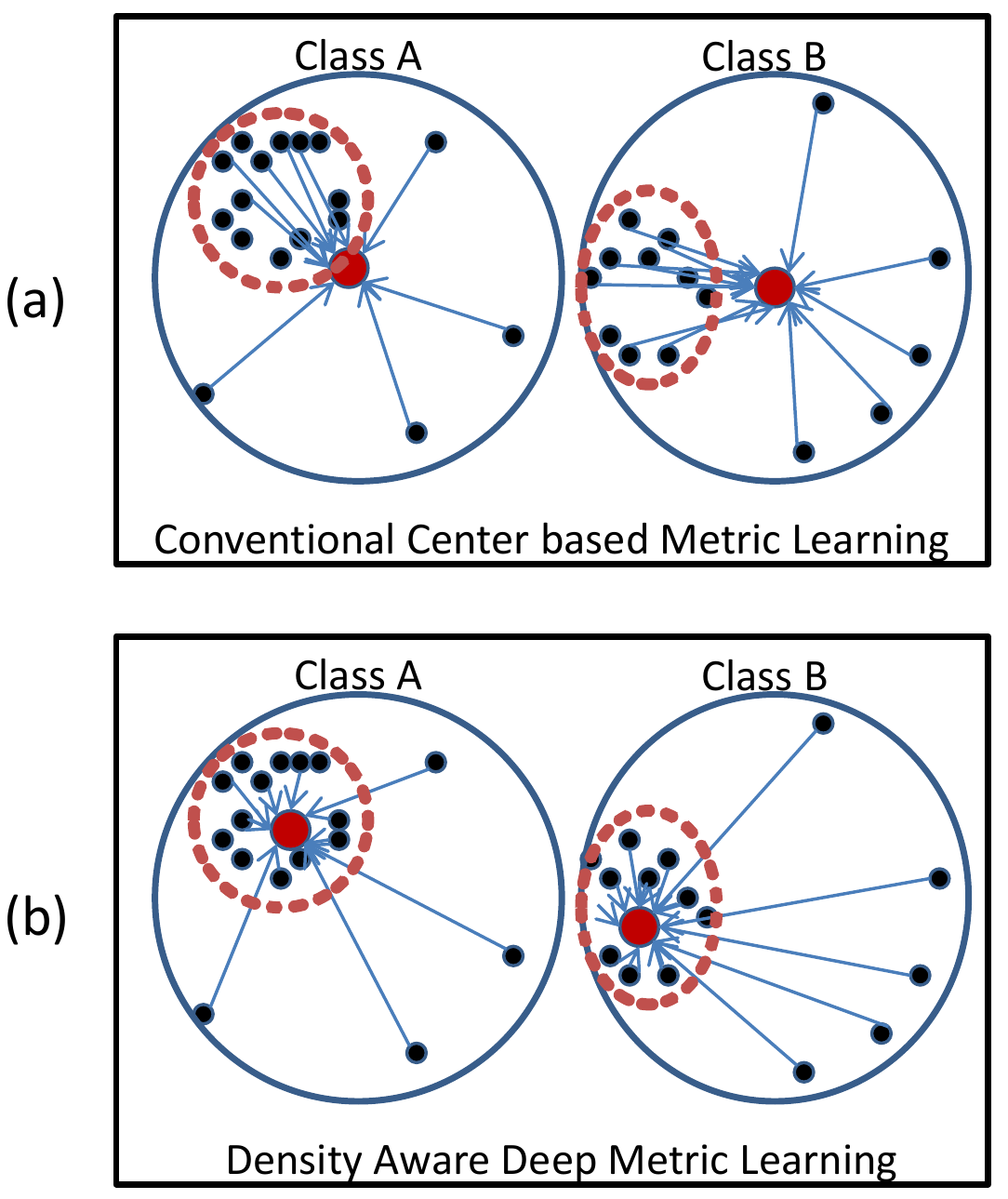}
     	\end{center}
     	\vspace{-5pt}
     	\caption{Illustrating the difference in the conventional and proposed metric learning techniques. (a) Conventional center loss based deep metric learning algorithms pull the data of a class towards the centroid of that class. (b) The proposed density aware deep metric learning algorithm pulls the samples of every class towards the most dense region of the respective clusters.}  
     	\label{graphical_abs}
        \vspace{-13pt} 
     \end{figure}


Classification models such as Convolutional Neural Networks (CNN) utilize deep metric learning based loss function for learning discriminative embeddings. The loss function attempts to bring the embeddings of the same classes close to each other in the output manifold. In this embedding space, a direct computation of the distance gives the dissimilarity score between the two images. Several different applications have investigated the use of deep metric learning algorithms such as person re-identification~\cite{cheng2016person,hermans2017defense,shi2016embedding}, 3D object retrieval~\cite{he2018triplet}, biometric recognition~\cite{btas18,schroff2015facenet,sun2014deep1,sun2014deep}, robot perception~\cite{kumar2016learning}, patch matching~\cite{han2015matchnet,zagoruyko2015learning}, and object recognition~\cite{masci2014descriptor,wohlhart2015learning}. 

In the literature, very efficient deep metric learning methods have been proposed such as triplet loss~\cite{schroff2015facenet} and quadruplet loss~\cite{chen2017beyond}. However, a major limitation of these loss functions is their heavy dependence on mining of hard samples for training~\cite{hermans2017defense,schroff2015facenet,shi2016embedding,yuan2017hard}. In the triplet Loss~\cite{schroff2015facenet}, for $N$ training classes and $K$ samples in each class, the total number of triplets for training can be as high as $N(N-1)K^2(K-1)$, which increases the training time on large datasets significantly. Another limitation of these methods is slow convergence, this heavily depends on the appropriate choice of the training curriculum. Further, the presence of outliers (noisy/poor-quality samples) in the training data, and their participation in triplets may hurt the training process. 
To the best of our knowledge there has been no study to understand the effect of outliers and density distribution of the training data on the performance of deep metric learning algorithms. 

As seen in Figure~\ref{graphical_abs}(a), conventional center loss based deep metric learning methods~\cite{he2018triplet,wen2016discriminative} generate embeddings of each class that lie closer to the centroid of the samples of that particular class. However, they do not take into account the distribution of the training data. In cases where outliers are present, the convergence of such methods on large databases can be slow and the outliers/noisy training samples can adversely affect the training of a discriminative model. In order to mitigate this challenge, the proposed algorithm minimizes the effect of outliers by calculating the center, taking into account the most dense region of the respective clusters for each class (Figure \ref{graphical_abs}(b)). Using the philosophy of the classical mean-shift algorithm~\cite{comaniciu2002mean}, the estimate of the mean is shifted to a denser region from the initial estimate of the centroid. This shifted center embedding is used for learning a discriminative model. The research contributions of the paper can be summarized as follows:
\begin{itemize}

\item The proposed density aware deep metric learning algorithm provides a generalized framework which can be augmented with any deep metric learning method for effective training especially with noisy data. 

\item Detailed analysis and comparison with other popular deep metric learning methods on four challenging databases pertaining to face and object images show that the proposed approach gives better recognition accuracies, exhibits superior convergence with reduced training time and is resilient to noisy training data.   
\end{itemize}

\section{Related Work}
Hadsell \etal \cite{hadsell2006dimensionality} proposed the contrastive loss, which was one of the first deep metric learning methods for training a discriminative model with a deep neural network. They used a single loss function to pull positive pairs and push negative pairs in the output embedding space of the model. This method of training a discriminative neural network, popularly known as the Siamese Network, resulted in several extensions~\cite{masci2014descriptor,sun2014deep1,sun2014deep} which produced excellent results on a variety of image recognition problems.  
Recently, one of the most popular methods for deep metric learning is the triplet loss~\cite{schroff2015facenet}. The triplet loss enforces the model to learn an embedding space where samples of similar classes are mapped closer to each other and that of other classes are pushed away. Wen \etal \cite{wen2016discriminative} used a combination of the softmax and the center loss for face recognition. Later, Chen \etal \cite{chen2017beyond} proposed the quadruplet loss which used an extra negative sample in addition to the anchor, positive and the negative sample that were utilized by the triplet loss. They showed that the extra negative term helps to train a more generalizable model. Thereafter, several methods have attempted to improve upon the triplet and the quadruplet loss based methods. Yuan \etal \cite{yuan2017hard} proposed an ensemble based technique for mining hard examples which are used for training a deep network using the contrastive loss. Hermans \etal \cite{hermans2017defense} proposed a triplet mining technique, by selecting the $k$ hardest positive samples and $k$ hardest negative samples for each anchor image in a batch of $N$ randomly sampled images from the training set. Recently, He \etal \cite{he2018triplet} proposed the triplet center loss where the center of the set of anchors and the center of the nearest negative cluster were utilized in the loss function of the triplet loss, for person re-identification.

\section{Density Aware Metric Learning}
The proposed method presents a novel contribution to the deep metric learning paradigm by incorporating the density of data in the clusters during training. Before delving into the detailed formulation, a brief illustration of the background is discussed.

\begin{figure*}[h!]
     	\begin{center}
     		\includegraphics[width=.95\linewidth]{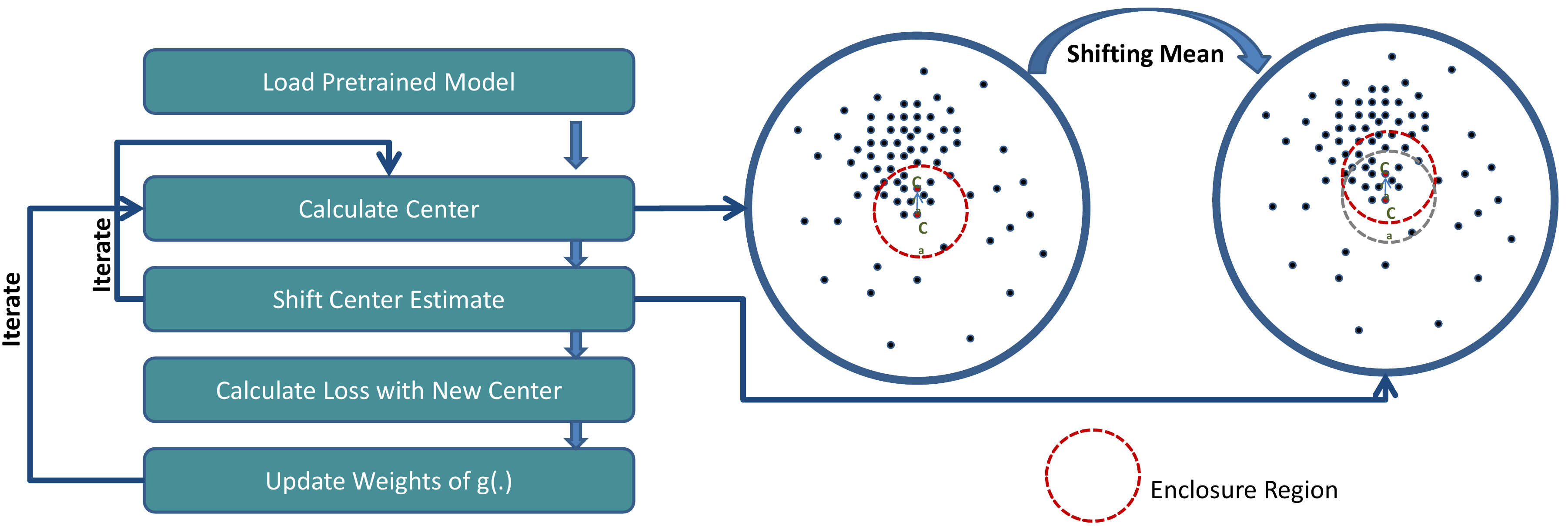}
     	\end{center}
         \vspace{-5pt}
     	\caption{The proposed algorithm iteratively finds the estimate for the center in the most dense region of the cluster. This center, when used with a deep metric learning algorithm, is expected to provide effective training and better convergence. (best viewed in color).}  
        \vspace{-2pt}
     	\label{pipeline}
     \end{figure*}

\subsection{Background}
\label{background}
In a classical pattern classification scenario, data $\vec{Z}$ from $n$ different classes is available, $\vec{Z}=\{z_1,z'_1,z_2,....,z_i,z'_i,...z_n\}$, where $z_i$ and $z'_i$ are two images of the same class $i$. Let the $i^{th}$ class contain $n_i$ number of training samples. The goal of a deep metric learning algorithm is to learn a function $g_\theta(z):\mathbb{R}^S \longrightarrow \mathbb{R}^T$ where $S$ is the dimensionality of the source data manifold, $T$ is the dimensionality of the output embedding space of the model $g$, and $\theta$ represents the trainable parameters of the model. 
For illustration, let $\{x,y\}$ be the pair of points on the embedding manifold of the model $g$. The distance metric function is defined as:
\begin{equation}
D\{x,y\}: \mathbb{R}^T \times \mathbb{R}^T \longrightarrow \mathbb{R}
\end{equation}
In this paper, Euclidean distance is used as the distance metric, which can be defined as:
\begin{equation}
D\{x,y\}= \left \|g_{\theta}(x) - g_{\theta}(y) \right \|^2_2
\end{equation}

Inspired from Large Margin Nearest Neighbor Classification~\cite{weinberger2009distance}, a typical deep metric learning loss $L$ can be used which is minimized by pulling intra-class embeddings together into one cluster and pushing the inter-class embeddings. From the training set $Z$, a 3-tuple is formed using three images, $z_a$, which is a sample of the class $a$, a positive sample $z'_a$, which is another image of the same class $a$, and a negative sample $z_b$ which is an image of another class $b$. The loss function can be expressed as:

   \begin{equation}
   \label{main_metric}
     L=  \left[D\{\vec{Z}_a,\vec{Z'}_a\} - D\{\vec{Z}_a,\vec{Z}_b\} +\alpha \right]_+ 
      \end{equation}
\begin{equation*}
     \forall (\vec{Z}_a, \vec{Z'}_a, \vec{Z}_b) \in \tau
\end{equation*}
where, $\tau$ is the set of all 3-tuples in the training data, $[f]_+=max(f,0)$, $\alpha$ is the margin parameter and $\vec{Z}_a$, $\vec{Z'}_a$ and $\vec{Z}_b$ are sets of all the anchors, positive and negative samples, prepared from the training set.  

   


\subsection{Proposed Formulation}
In order to present the proposed approach, the standard loss metric (Equation~\ref{main_metric}) is re-formulated where the anchor $z_a$ is replaced with the center of class $a$. The center embedding $C_a$ is calculated as the mean of all the embeddings of class $a$. Thus, the loss function may be expressed as:
\begin{equation}
   \label{modified_metric}
     L=  \left[D\{C_a,\vec{Z'}_a\} - D\{C_a,\vec{Z}_b\} +\alpha \right]_+ 
      \end{equation}
\begin{equation*}
 where \quad \quad C_a=\frac{\sum_{n_a} g(z_{a})}{n_a}
\end{equation*}

$C_a$ represents the centroid of the cluster corresponding to class $a$, containing $n_a$ training samples. However, depending on the density of the cluster (as shown in Figure~\ref{pipeline}), the mean-shift algorithm~\cite{comaniciu2002mean} may be applied to iteratively arrive at the mean driven by non-parametric density estimation of the cluster. 

\subsubsection{Shifting the Mean to a Denser Region}
\label{mean_shift}

Reiterating, $C_a$ is the initial centroid of the cluster which is calculated by taking the mean of all the embeddings of  class $a$. Now, from $C_a$, selecting the nearest $p$ points (in the embedding manifold of the model $g$) out of all the $n_a$ points in the cluster corresponding to class $a$, we take the mean of only these $p$ points, where $p < n_a$. We term the region of the embedding manifold containing these $p$ points around the centroid (the red dotted circle in Figure~\ref{pipeline}) as the \textbf{enclosure region}, and the set of these $p$ points as \textbf{enclosure points}. The estimate for the new center $C'_a$ can be calculated as,  

\begin{equation}
C'_a=\frac{\sum_{i=1}^p g(z_{a}^{i})}{p} \quad \forall z_{a}^{i} \in \{ z_{a}^{1}, z_{a}^{2}...z_{a}^{p} \}
\end{equation}
where, $z_{a}^{i}$ is the $i^{th}$ point inside the enclosure region. Figure~\ref{pipeline} shows the new mean $C'_a$, which is expected to be in a denser region of the cluster. The difference of the new mean $C'_a$ and the old mean $C_a$ gives the mean shift vector which can be expressed as:
\begin{equation}
V_a=\left[ \frac{\sum_{i=1}^p g(z_{a}^{i})}{p} - \frac{\sum_{n_a} g(z_{a})}{n_a} \right] 
\end{equation}
This process is repeated iteratively until the mean shift is negligible, thus leading to convergence. 


\subsubsection{Weighted Mean Shift} The above calculation of the centroid does not take into account any weightage of the points around the mean that are considered. In order to give importance to the points nearer to the centroid, we can use a weight coefficient $W_i$ for every point $i$ in the \emph{enclosure region}.
The $k^{th}$ estimate of the center with respect to weights $W_i$ can be calculated as,

\begin{equation}
 C^k_a=\frac{\sum_{i=1}^p W_i C^{k-1}_a g(z_{a}^{i})}{\sum_{i=1}^p W_i C^{k-1}_a} \quad \forall z_{a}^{i} \in \{    z_{a}^{1}, z_{a}^{2}...z_{a}^{p} \}
\end{equation}
$C^{k-1}_a$ being the $(k-1)^{th}$ estimate of the mean. The corresponding mean shift vector may be expressed as,
\begin{equation}
V^k_a=\left[ \frac{\sum_{i=1}^p W_i C^{k-1}_a g(z_{a}^{i})}{\sum_{i=1}^p W_i C^{k-1}_a} - C^{k-1}_a \right]
\end{equation}
Here, $p$ is the number of \emph{enclosure points} for the $k^{th}$ iteration, and $z_{a}^{i}$ is the $i^{th}$ data point for the class $a$.

\subsection{Selecting weights using a Kernel Density Estimate (KDE)}    
In order to select weights $W_i$ for each point $i$ in the cluster represented by the centroid $C_a$ for a particular class $a$, we can use a kernel density estimate that are generally used by non-parametric density estimation techniques. A uniform kernel for selecting the weights can be expressed as:
\begin{equation}
W_i = \left\{\begin{matrix}
 c \quad if \left \| C_a - z^i_a \right \| < f& \\ 
 0 \quad \quad \quad \quad otherwise & 
\end{matrix}\right.
\end{equation}
where, $\left \|C_a - z^i_a\right \|$ gives the distance of the point $z^i_a$ from the cluster centroid $C_a$ for class $a$. The uniform kernel assigns a weight $c$ to the point $z^i_a$ if it is within the \emph{enclosure region}. The \emph{enclosure region} has a radius of $f$, thus all the points which are at a distance of $f$ or less from the centroid $C_a$ are assigned the same weight $c$. Instead of directly using a parameter for the radius of the \emph{enclosure region}, $p$ nearest \emph{enclosure points} can also be considered out of all the points in the cluster for class $a$. Algorithm~\ref{algo} outline the steps of the proposed approach using the triplet loss.

\begin{algorithm}[h!]
	\SetAlgoLined
	\small
	\KwIn{CNN model $g_\theta$, training data \{{$\vec{Z}$\} } }
	
	\KwOut{Trained model $g_\theta$}
	\textbf{Parameters:} $e$ (epochs), $\theta$ (parameters of $g$), $m$ (batch size), $k$ (number of batches) $p$ (number of enclosure points) $s$ (mean shift iterations), $t^p$ (threshold for hard positive selection), $t^n$ (threshold for hard negative selection), $f$ (radius of enclosure region) \\
	
	\nl \For{Epoch=1 to $e$}{
		 \textbf{Generate Triplets:} 
		
         \nl \textbf{Initialize:} $X=\{\}$ (empty set of selected Hard Triplets) 
        
        \nl \textbf{Initialize:} $Pool=\{\}$ (empty pool of samples) 

		\nl \For{every class $a=1$ to $n$}{
			
		\nl	Select b images randomly from class $a$ \\
        \nl	$Pool=Pool\cup$ selected images
         }      
        \nl \For{each image $z_{a}^{i}$ of each class $a$ in $Pool$}{
        \nl Select $z_{a}^{i}$ as the anchor image \\
			\nl \For{each image $z_{l}^{y}$ in $Pool$ such that $z_{l}^{y}\neq z_{a}^{i}$}{
		\nl if $a=l$ and $D\{z_{a}^{i},  z_{l}^{y}\}  > t^p$ then $X=X\cup  z_{l}^{y}$\\
        \nl if $a\neq l$ and $D\{z_{a}^{i},  z_{l}^{y}\}  < t^n$ then $X=X\cup  z_{l}^{y}$

            }
	}

  \textbf{Calculate Center:} \\
        \nl $C_a=\frac{\sum_{n_a} g(z_{a})}{n_a}$
        
        \textbf{Shift Center:} \\
         
          \nl \For{every class $a$}{
         
         \nl \For{k=1 to s}{
         
       \nl  $W_i=K_u(C^{k-1}_a - z^i_a) = \left\{\begin{matrix}
 c \quad if \left \| C^{k-1}_a - z^i_a \right \| < f& \\ 
 0 \quad \quad \quad \quad otherwise & 
\end{matrix}\right. $
         
        \nl $ C^k_a=\frac{\sum_{i=1}^p W_i C^{k-1}_a g(z_{a}^{i})}{\sum_{i=1}^p W_i C^{k-1}_a} \quad \forall z_{a}^{i} \in \{    z_{a}^{1}, z_{a}^{2}...z_{a}^{p} \} $
         
       \nl  $V^k_a=\left[ \frac{\sum_{i=1}^p W_i C^{k-1}_a g(z_{a}^{i})}{\sum_{i=1}^p W_i C^{k-1}_a} - C^{k-1}_a \right] $
         
         } }

      \textbf{Generate embeddings}\\
		\nl \For{every batch of size $m$}{

			\nl Forward pass through $g$ to find $g_\theta(Z_{a}),g_\theta(Z'_{a}), f_\theta(Z_{b})$ 
			
			 \textbf{Calculate loss $L$} \\
			
          \nl $ L=  \sum_{m}\left [\left \|   C^s_a - g(\vec{Z'}_a)\right \| ^2_2 - \left \|    C^s_a - g(\vec{Z}_b)\right \| ^2_2 + \alpha \right]  $

			 \textbf{Calculate gradient} \\
		\nl	$\bigtriangleup W=\nabla_{\theta} \frac{1}{m}\sum_{m} L$
			
			\nl  \textbf{Update weights} of $g_\theta$ using $\bigtriangleup W$ 	
		}

    }
	
	\caption{Density Aware Triplet Loss.}
    \label{algo}
\end{algorithm}

\section{Density Aware Deep Metric Learning in Triplet and Quadruplet Loss}
The proposed Density Aware Metric Learning is a generic formulation and can be incorporated into any deep metric learning loss function. Here, we present the formulations of triplet and quadruplet loss based density aware metric learning: 

\subsection{ Density Aware Triplet Loss (DATL)}
Schroff \etal ~\cite{schroff2015facenet} proposed the triplet loss based deep metric learning technique where the loss $L$ is minimized by the same philosophy as discussed in Section~\ref{background}. From the training set $Z$, a triplet is formed using an anchor $z_a$, which is an image of the class $a$, a positive sample $z'_a$, which is another image of the same class $a$, and a negative sample $z_b$ which is an image of another class $b$. The loss function is expressed as, 
\vspace{6pt}
\begin{equation}
     L= \left[\left \|   g(\vec{Z}_a) - g(\vec{Z'}_a)\right \| ^2_2 - \left \|   g(\vec{Z}_a) - g(\vec{Z}_b)\right \| ^2_2 + \alpha \right]_+
      \end{equation}
   \begin{equation*}
     \forall (\vec{Z}_a, \vec{Z'}_a, \vec{Z}_b) \in \tau
\end{equation*}
where $\vec{Z}_a$, $\vec{Z'}_a$ and $\vec{Z}_b$ are sets of all anchors, positive and negative samples, respectively, and $\tau$ is the set of all triplets in the training data. Using the proposed approach, the anchor is replaced with the center which is iteratively determined ($C_a$ or $C'_a$, and so on) with an appropriate kernel density estimate. The loss function for the Density Aware Triplet Loss (DATL) is as follows:
\begin{equation}
\label{triplet_center}
    L= \left[ \left \|   C_a - g(\vec{Z'}_a)\right \| ^2_2 - \left \|   C_a - g(\vec{Z}_b)\right \| ^2_2 + \alpha \right]_+
\end{equation}
\subsection{ Density Aware Quadruplet Loss (DAQL)}
The triplet loss is extended by Chen \etal \cite{chen2017beyond} as the quadruplet loss where a second negative image $z_c$ is introduced. The loss function for the same in the proposed density aware paradigm can be expressed as,
\begin{multline}
     L= \left [\left \|   C_a - g(\vec{Z'}_a)\right \| ^2_2 - \left \|   C_a - g(\vec{Z}_b)\right \| ^2_2 + \alpha_1 \right]_+ \\ + \left[ \left \|  C_a - g(\vec{Z'}_a)\right \| ^2_2 - \left \|   C_a - g(\vec{Z}_c)\right \| ^2_2 + \alpha_2 \right]_+
\end{multline}
\begin{equation*}
     \forall (\vec{Z}_a, \vec{Z'}_a, \vec{Z}_b, \vec{Z}_c) \in \vartheta
\end{equation*}
where $\vartheta$ is the set of all the quadruplets prepared from the training set. 

\subsection{Experimental Setup and Implementation}
The deep CNN architecture by Wu \etal \cite{wu2018light} is utilized to learn a discriminative model with the proposed loss function. The weights are initialized from a network that is pretrained on the MS-Celeb 1M dataset. 
The model has $17$ convolutional layers, along with 10 Max-Feature-Map layers. The network has two fully connected layers at the end, producing embeddings of dimensionality $256$. Training is performed using the Adam optimizer. The batch size is kept at $60$ and the learning rate of $10^{-3}$ is used which is decreased gradually till $10^{-7}$. Hard mining is performed (steps 7-11 of Algorithm~\ref{algo}) for all the variants of triplet and quadruplet losses according to the \emph{Batch Hard} scheme proposed by Hermans \etal \cite{hermans2017defense}. The hard mining is only performed at the end of training (once the learning
plateaus) to accelerate the training process. All the codes are implemented using the Pytorch platform on a machine with Intel Core i7 CPU, 64GB RAM and NVIDIA GTX 1080Ti GPU.

\section{Experiments}
The proposed algorithm is evaluated on the SCface~\cite{grgic2011scface} and FaceSurv~\cite{fgcscrv} datasets for cross-modal face matching, and on the CIFAR10~\cite{krizhevsky2009learning} and STL-10~\cite{coates2011analysis} datasets for object recognition.


\subsection{Datasets}
Details of the databases and experimental protocols are described in this section. 

\noindent\textbf{SCface~\cite{grgic2011scface}} is a face dataset containing poor quality face images captured from surveillance cameras in indoor environment. The database contains 4160 images of 130 subjects. The images are captured with eight different cameras, out of which two cameras operated in night-vision mode and one camera is operated in Near-Infrared mode. The images are taken from three different stand-off distances namely 4.2 mts, 2.6 mts, and 1 mt. Out of the 130 subjects, images of 50 subjects are used for training and the remaining are used for testing. The classes/subjects in the train and test set are non overlapping.\\
\textbf{FaceSurv~\cite{fgcscrv}} is a video face database where the subjects walk towards the camera from a distance of about 10 mts. It has 396 daytime and 365 nighttime videos of 240 subjects. The nighttime videos are captured in complete darkness with an NIR illuminator. Each video has about 200 frames on an average. Each subject has three gallery images which are captured in controlled scenarios from a standoff distance of 1 mt. Videos of only 39 subjects are used for training and the remaining are used for testing. The classes/subjects in train and test sets are disjoint.\\
\textbf{CIFAR-10~\cite{krizhevsky2009learning}} is a popular object recognition dataset consisting 60,000 images of 10 classes.
The resolution of the images is $32 \times 32$.
The training set contains 50,000 images (5,000 images of each class) and 10,000 images (1,000 per class) comprise the testing set. \\
\textbf{STL-10~\cite{coates2011analysis}} contains 113,000 images of 10 different objects. The resolution of the images is $96 \times 96$. The total number of images for training and testing are 5,000 (500 per class) and 8,000 (800 per class), respectively. The remaining images are unlabeled and have not been used for the experiments.

\subsection{Evaluation Criteria}
In this work, two different kinds of experiments have been performed: cross-modal face recognition (identification) and object recognition (retrieval). 
For face identification, the test set is partitioned into probe (query images) and gallery (reference set/database) sets. For every image of the probe set, matching is performed with each image of the gallery by a forward pass through the learned model ($g_\theta$), followed by computing Euclidean distance between the embeddings of the probe and the gallery images to calculate the match score. Rank $\mathcal{K}$ accuracy is the ratio (multiplied by $100$ to get a percentage) of the number of times the correct class is among the top $\mathcal{K}$ matches to the number of matching attempts (once for each probe image). Recall @ $\mathcal{K}$ is the average recall score for all the query images. Following the definition by Song \etal \cite{song2017deep}, the recall score is one if the relevant class is retrieved in the top $\mathcal{K}$ matches with the gallery/database set, and is zero otherwise.         

\section{Results}
The experiments have been performed by partitioning each database into the train and test sets. The CNN model is trained on the train set using the proposed density aware deep metric learning, i.e. the Density Aware Triplet Loss (DATL) and the Density Aware Quadruplet Loss (DAQL). Comparisons have been performed with the vanilla triplet and quadruplet losses, (their variants for cross-modal matching are implemented for the SCface and FaceSurv databases). In addition, the proposed algorithm is also compared with hard triplet loss~\cite{hermans2017defense} and recently proposed triplet center loss~\cite{he2018triplet}. The former is a variant of the vanilla triplet loss using a moderate hard mining approach. Triplet center loss is a formulation which mimics the conventional center based triplet loss previously discussed (Equation~\ref{modified_metric}).
\begin{table}[]
\caption{Summarizing the results of face identification on the SCface~\cite{grgic2011scface} Database.}
\scalebox{.91}{
\begin{tabular}{|l|l|l|l|l|}
\hline
\multicolumn{2}{|l|}{\multirow{2}{*}{\textbf{Method}}}         & \multicolumn{3}{c|}{\textbf{Identification (\%) (Rank 1)}}         \\ \cline{3-5} 
\multicolumn{2}{|l|}{}                   & \multicolumn{1}{c|}{\textbf{24 x 24}} & \textbf{32 x 32} & \textbf{  48 x 48} \\ \hline
\multicolumn{2}{|l|}{MDS~\cite{biswas2013pose}} &          64.87               & 70.48            & 76.14            \\ \hline
\multicolumn{2}{|l|}{Co-Transfer Learning~~\cite{bhatt2014improving}}  &        70.14                 & 76.29            &  83.47           \\ \hline
\multicolumn{2}{|l|}{Res-Net~\cite{wen2016discriminative}}            &          36.30               & 81.80            &          94.30   \\ \hline
\multicolumn{2}{|l|}{Coupled Res-Net~\cite{lu2018deep}}               &           73.30               & 93.50            &        98.00    \\ \hline
\multicolumn{2}{|l|}{VGGFace~\cite{lu2018deep}}                       &        41.30                 & 75.50            &         88.80    \\ \hline
\multicolumn{2}{|l|}{Coupled VGGFace~\cite{lu2018deep}}               &       62.30                  & 91.00            &       94.80      \\ \hline
\multicolumn{2}{|l|}{Coupled Light-CNN~\cite{lu2018deep}}             &          50.50               & 85.00            &         94.00    \\ \hline
\multicolumn{2}{|l|}{Triplet loss~\cite{liu2016transferring}}         &       70.69                  & 95.42            &        97.02     \\ \hline
\multicolumn{2}{|l|}{Quadruplet loss~\cite{chen2017beyond}}           &        74.00                 & 96.57            &      98.41       \\ \hline
\multicolumn{2}{|l|}{Hard triplet loss~\cite{hermans2017defense}}     &      72.65                   & 96.12            &       98.05      \\ \hline
\multicolumn{2}{|l|}{Triplet Center Loss~\cite{he2018triplet}}        &        75.45                 & 96.10            &      \textbf{98.50}      \\ \hline
\multicolumn{2}{|l|}{Discriminative MDS~\cite{yang2018discriminative}}        &        62.70                 & 65.50            &      70.70       \\ \hline
\multirow{2}{*}{Proposed}                & DATL                & \textbf{76.24 }              & \textbf{96.87}   &  98.09   \\ \cline{2-5} 
                                         & DAQL                & \textbf{77.25}               & \textbf{96.58}   &  98.14  \\ \hline
\end{tabular}}
\label{tab:scf}
\end{table}
\begin{table}[]
\caption{Summarizing the results of face identification on the FaceSurv~\cite{fgcscrv} database.}
\scalebox{.92}{
\begin{tabular}{|l|l|c|c|c|}
\hline
\multicolumn{2}{|l|}{\multirow{2}{*}{\textbf{Method}}}     & \multicolumn{3}{c|}{\textbf{Identification (\%) (Rank 1)}}                                                      \\ \cline{3-5} 
\multicolumn{2}{|l|}{}     & \textbf{24 x 24}                   & \textbf{32 x 32}                   & \multicolumn{1}{l|}{\textbf{  48 x 48}} \\ \hline
\multicolumn{2}{|l|}{Triplet loss~\cite{liu2016transferring}}     &   18.0                             & 38.5                               &    72.5                               \\ \hline
\multicolumn{2}{|l|}{Quadruplet loss~\cite{chen2017beyond}}       &     16.4                            & 38.5                               &    77.9                              \\ \hline
\multicolumn{2}{|l|}{Hard triplet loss~\cite{hermans2017defense}} &    17.8                             & 40.8                               &    78.9                              \\ \hline
\multicolumn{2}{|l|}{Triplet Center Loss~\cite{he2018triplet}}    &     18.4                           & 41.5                               &        82.7                           \\ \hline
\multirow{2}{*}{Proposed}  & DATL    & {\textbf{20.4}} & {\textbf{42.8}} & {\textbf{ 85.9}}    \\ \cline{2-5} 
                           & DAQL & {\textbf{21.3}} & {\textbf{48.6}} & {\textbf{  85.6}}    \\ \hline
\end{tabular}}
\label{tab:cscrv}
\end{table}
\begin{table}[!t]
\caption{Summarizing the results of object retrieval on the CIFAR-10~\cite{krizhevsky2009learning} database.}
\scalebox{0.95}{
\begin{tabular}{|l|l|l|l|l|}
\hline
\multicolumn{2}{|l|}{\multirow{2}{*}{\textbf{Method}}}     & \multicolumn{3}{c|}{\textbf{Recall @ $\mathcal{K}$ (\%)}}                            \\ \cline{3-5} 
\multicolumn{2}{|l|}{}     & \multicolumn{1}{c|}{\textbf{$\mathcal{K}$ = 1}} & \textbf{$\mathcal{K}$ = 10} & \textbf{$\mathcal{K}$ = 100} \\ \hline
\multicolumn{2}{|l|}{Triplet loss~\cite{liu2016transferring}}     & 76.24                               & 94.78           & 97.21            \\ \hline
\multicolumn{2}{|l|}{Quadruplet loss~\cite{chen2017beyond}}       & 78.35                               & 95.40           & \textbf{98.99}            \\ \hline
\multicolumn{2}{|l|}{Siamese+Triplet~\cite{kumar2016learning}} & 78.62                               & 92.57           & 97.19            \\ \hline

\multicolumn{2}{|l|}{Hard triplet loss~\cite{hermans2017defense}} & 78.51                               & 93.87           & 97.41            \\ \hline
\multicolumn{2}{|l|}{Triplet Center Loss~\cite{he2018triplet}}    & 79.41                               & 96.10           & 95.78            \\ \hline
\multirow{2}{*}{Proposed}  & DATL    & \textbf{80.34}                      & \textbf{96.68 } & 97.84   \\ \cline{2-5} 
                           & DAQL & \textbf{80.81 }                     & \textbf{96.12 } & 97.58   \\ \hline
\end{tabular}}
\label{tab:cifar10}
\end{table}
\begin{table}[]
\caption{Summarizing the results of object retrieval on the STL-10~\cite{coates2011analysis} database.}
\scalebox{1}{
\begin{tabular}{|l|l|l|l|l|}
\hline
\multicolumn{2}{|l|}{\multirow{2}{*}{\textbf{Method}}}     & \multicolumn{3}{c|}{\textbf{Recall @ $\mathcal{K}$ (\%)}}                            \\ \cline{3-5} 
\multicolumn{2}{|l|}{}     & \multicolumn{1}{c|}{\textbf{$\mathcal{K}$ = 1}} & \textbf{$\mathcal{K}$ = 10} & \textbf{$\mathcal{K}$ = 100} \\ \hline
\multicolumn{2}{|l|}{Triplet loss~\cite{liu2016transferring}}     & 72.47                               & 78.54           & 80.41            \\ \hline
\multicolumn{2}{|l|}{Quadruplet loss~\cite{chen2017beyond}}       & 73.98                               & 78.71           & 81.77            \\ \hline

\multicolumn{2}{|l|}{Siamese+Triplet~\cite{kumar2016learning}}       & 73.62                               & 77.15           & 81.34            \\ \hline

\multicolumn{2}{|l|}{Hard triplet loss~\cite{hermans2017defense}} & 72.95                               & 76.08           & 81.90            \\ \hline
\multicolumn{2}{|l|}{Triplet Center Loss~\cite{he2018triplet}}    & 74.61                               & 77.98           & 81.59            \\ \hline
\multirow{2}{*}{Proposed}  & DATL    & \textbf{75.27}                      & \textbf{79.08}  & \textbf{82.38}   \\ \cline{2-5} 
                           & DAQL & \textbf{75.84}                      & \textbf{80.17}  & \textbf{83.74}   \\ \hline
\end{tabular}}
\label{tab:stl10}
\end{table}

For face recognition, Rank 1 accuracies for three different probe resolutions, namely $48 \times 48$, $32 \times 32$ and $24\times24$ are reported. As shown in table \ref{tab:scf} on $32 \times 32$ and $24\times24$ resolutions, the proposed algorithm produces state-of-the-art results for the SCface database. It outperforms the vanilla triplet and the quadruplet losses and their variants on the FaceSurv database (Table \ref{tab:cscrv}) as well. Moreover, on the SCface database, we report published results from different cross-modal face recognition methods. As shown in Table \ref{tab:scf}, it can be observed that the proposed algorithm outperforms these existing algorithms, specifically for lower resolution levels.   

For the object retrieval task, experiments are performed on the CIFAR-10 and STL-10 datasets. As shown in Table \ref{tab:cifar10}, on the CIFAR-10 dataset, the proposed algorithm outperforms both the triplet and the quadruplet losses and their variants for recall @ 1 and recall @ 10. However, for recall @ 100 it produces competitive accuracy with respect to the other algorithms. As shown in Table \ref{tab:stl10}, on the STL-10 dataset, the proposed algorithm outperforms the vanilla triplet and quadruplet losses along with their variants on recall @1, 10 and 100. 

\section{Analysis and Discussion}
This section analyzes the performance of the proposed algorithm with respect to training with noisy data, convergence, training time, and parameters.  

\subsection{Effect of Noisy Data during Training}
\begin{figure}[t]
\centering
  \subfloat[]{\includegraphics[width=.96\linewidth]{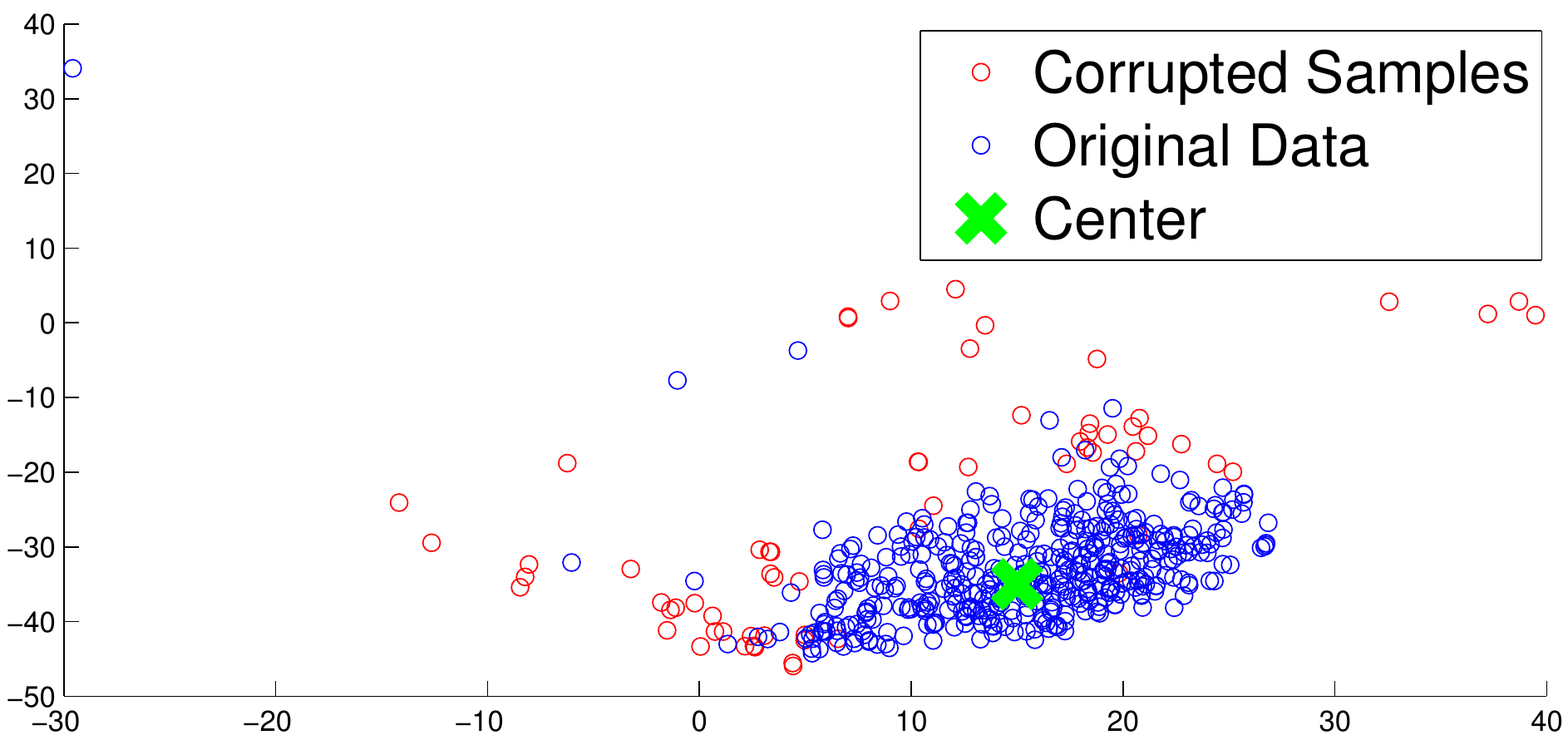}\label{noisy_tsne}}
  \hspace{0.001cm} 
  \subfloat[]{\includegraphics[width=.96\linewidth]{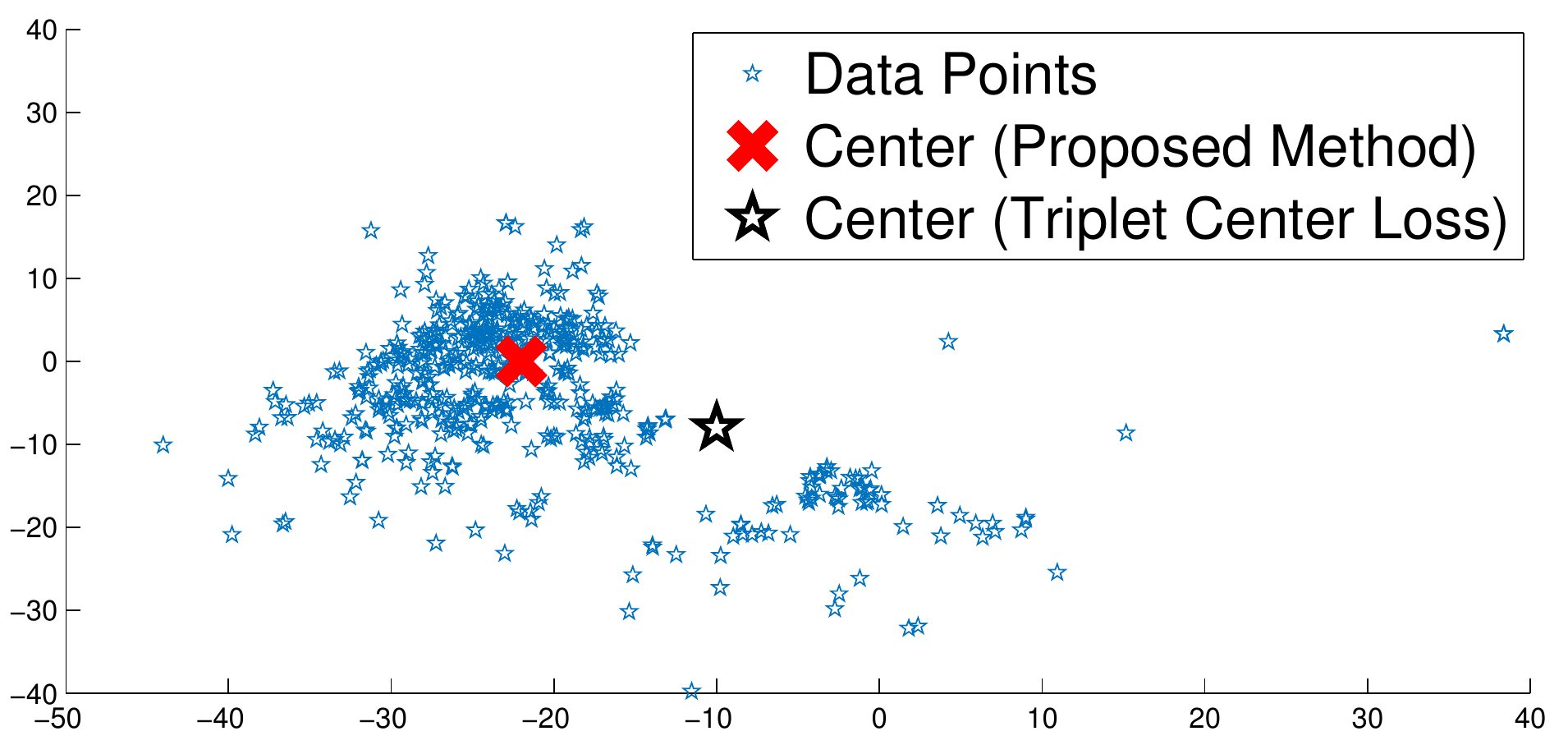}\label{noisy_center}}

  \vspace{0.001cm}
    \vspace{-10pt}
   \caption{ (a) tSNE Visualization of noisy samples (for a particular class), which shows that most of the noisy samples are outliers (b) Center computed by the proposed method is in the dense region of the class while the conventional center is away from the dense region. Visualization is on the STL-10 database, one particular class is shown for illustrative brevity (best viewed in color).}
  \vspace{-10pt}
   \label{fig:outlier}
    
   \end{figure}

One of the primary properties of the proposed method is the ability to ignore outliers during training. Such outliers may often be represented by noisy data (low resolution/quality and poor illumination). As shown in Figure~\ref{noisy_tsne}, these noisy data samples affect the training process of conventional deep metric learning based algorithms. Since the proposed method computes the cluster center only by using the points inside the enclosure region, the outliers are effectively ignored. On the other hand, conventional deep metric learning algorithms would consider all the points (including outliers) which may lead to unnecessary jitter in the convergence during training. 
An experiment is performed on the STL-10 database by replacing 15\% of samples from each class by low resolution variants ($32 \times 32$ and $24 \times 24$ as two separate experiments) of the same. Such training samples are expected to be outliers and thus may have potential to hurt the training process. We use a no reference image quality score (BRISQUE~\cite{mittal2012no}) (a lower score implies better image quality) for the original training samples of STL-10 which is 33.90 (average for training set). For the noisy samples, the score is 45.14 (for $24\times24$) and 41.83 (for $32\times32$). This infers that the low resolution data are of lower quality. As shown in Table~\ref{noisy}, the proposed methods perform better than conventional deep metric learning techniques when noisy data is introduced in the training process. It also exhibits that performance improvement is greater for the experiment where higher amount of data corruption (adding $24 \times 24$ images) is performed.

\begin{table}[h!]
\caption{Results on the STL-10 database after adding noisy training data for every class.}
\label{noisy}
\scalebox{.94}{
\begin{tabular}{|l|l|l|l|l|l|}
\hline
\multicolumn{2}{|l|}{\multirow{4}{*}{\textbf{Method}}} & \multicolumn{4}{l|}{\textbf{Resolution of Noisy Samples}}                     \\ \cline{3-6} 
\multicolumn{2}{|l|}{}                                 & \multicolumn{2}{c|}{\textbf{24 x 24}} & \multicolumn{2}{c|}{\textbf{32 x 32}} \\ \cline{3-6} 
\multicolumn{2}{|l|}{}                                 & \multicolumn{4}{c|}{\textbf{Recall @ $\mathcal{K}$ (\%)}}                                      \\ \cline{3-6} 
\multicolumn{2}{|l|}{}                                 & \textbf{$\mathcal{K}$=1}      & \textbf{$\mathcal{K}$=10}     & \textbf{$\mathcal{K}$=1}      & \textbf{$\mathcal{K}$=10}     \\ \hline
\multicolumn{2}{|l|}{Triplet Loss~\cite{schroff2015facenet}}                     & 54.12             & 58.00             & 68.45             & 72.58             \\ \hline
\multicolumn{2}{|l|}{Quadruplet Loss~\cite{chen2017beyond}}                  & 56.51             & 59.77             & 68.74             & 73.01             \\ \hline
\multicolumn{2}{|l|}{Hard Triplet Loss~\cite{hermans2017defense}}                & 58.29             & 60.41             &     65.37              &     72.91              \\ \hline
\multicolumn{2}{|l|}{Triplet Center Loss~\cite{he2018triplet}}              & 62.76             & 65.40             & 68.76             & 73.16             \\ \hline
\multirow{2}{*}{Proposed}            & DATL            & \textbf{66.52}    & 69.45             & \textbf{69.87}    & 73.21             \\ \cline{2-6} 
                                     & DAQL            & 65.47             & \textbf{69.80}    & 68.52             & \textbf{75.40}    \\ \hline
\end{tabular}}
\end{table}

\subsection{Size of the Enclosure Region}
One important parameter of the proposed algorithm is the size of the enclosure region.
For implementation, the enclosure region is determined by taking the nearest k\% points from the current center embedding point. A region of 20\% signifies that the nearest 20\% points (with respect to all the points of the particular cluster) from the current center are considered to be inside the enclosure region. 
\begin{figure}[t]
     	\begin{center}
     		\includegraphics[width=1\linewidth]{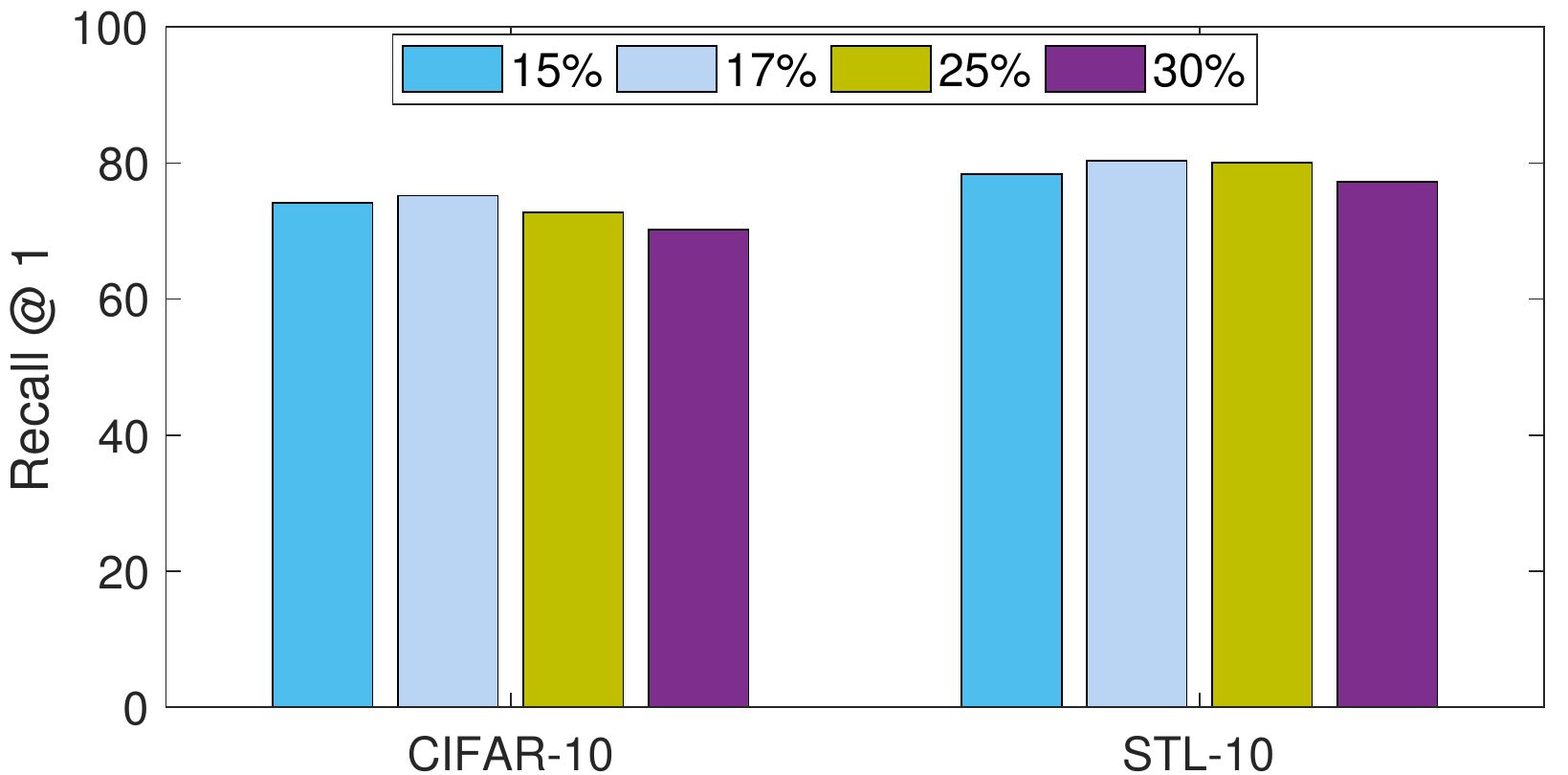}
     	\end{center}
         \vspace{-5pt} 
     	\caption{Effect of the size of the enclosure region on the proposed Density Aware Triplet Loss on the STL-10 and CIFAR-10 databases (best viewed in color).}  
     	\label{enclosure}
      
     \end{figure}
Figure~\ref{enclosure} shows the results for the 
proposed Density Aware Triplet Loss on the STL-10 and CIFAR-10 databases for four different enclosure regions.
It can be seen that an enclosure region of 17\% yields optimal results while larger or smaller enclosure region results in reduced accuracy on both the databases.   
\begin{figure}[!t]
\centering
  \subfloat[]{\includegraphics[width=4.1cm]{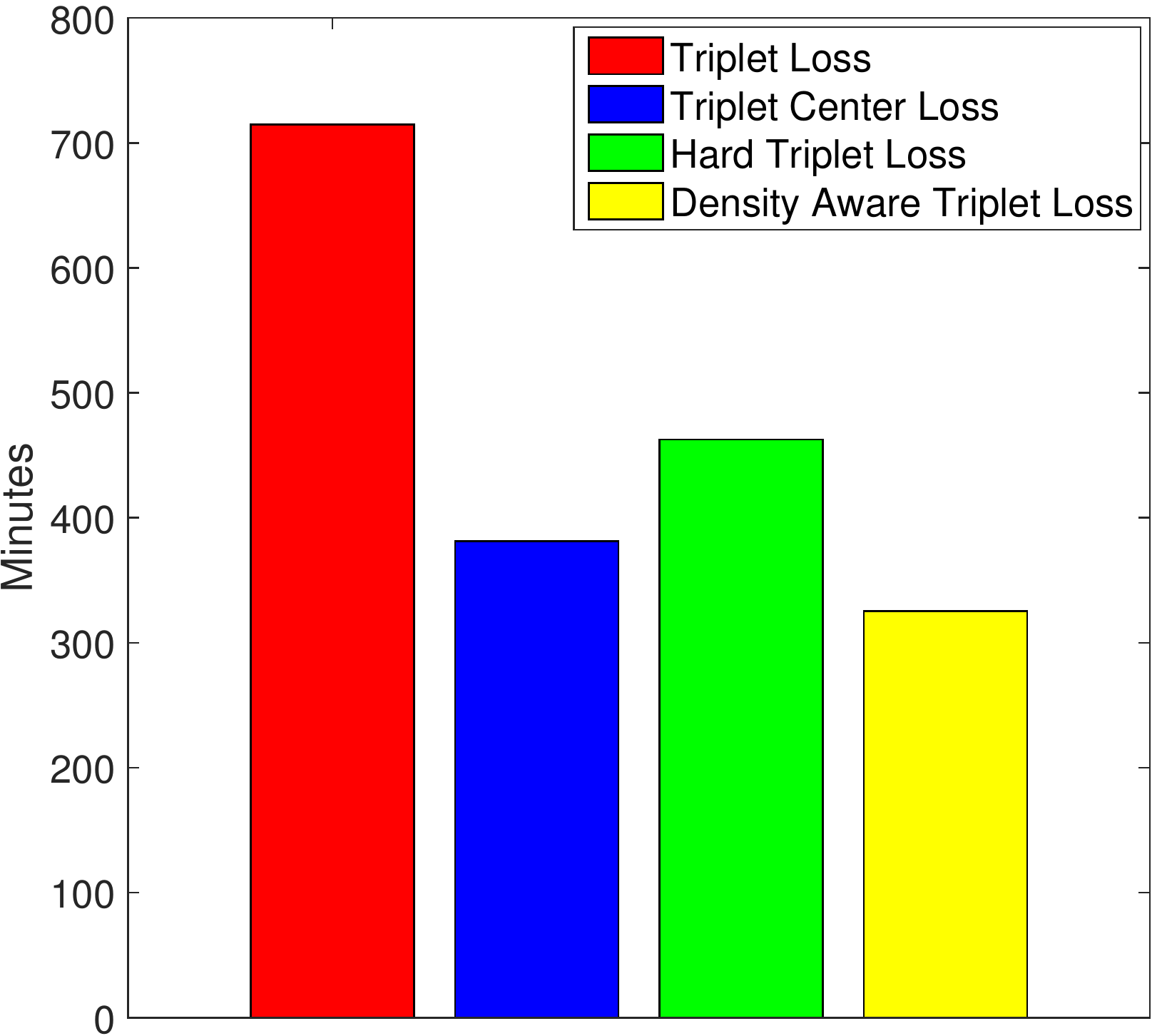}\label{comp_scf}}
  \hspace{0.001cm} 
  \subfloat[]{\includegraphics[width=4.05cm]{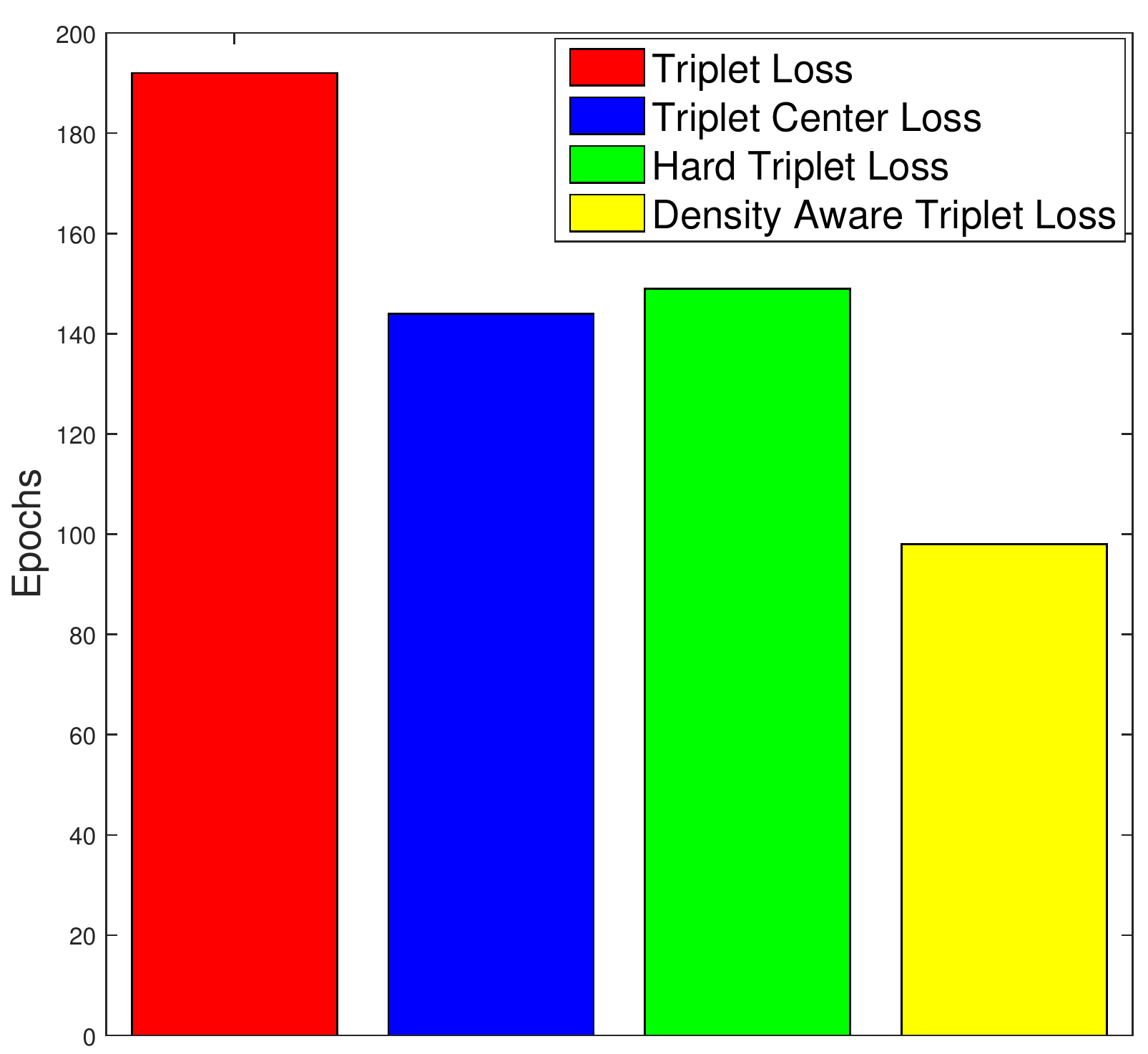}\label{comp_casia}}
  

   \caption{Total (a) training time and (b) epochs for training on the STL-10 database (best viewed in color).}
  
   \label{fig:time}
    
   \end{figure}
\subsection{Training Time}

Owing to better convergence properties of the proposed DATL, total training time required is much less as compared to the vanilla triplet loss and its variants. As shown in Figure \ref{fig:time}(a), 
the total time needed to train the proposed algorithm is 325.4 minutes. On the other hand, the vanilla triplet loss, triplet center loss, and the hard triplet loss requires 714.9, 381.4, and 462.7 minutes, respectively on the STL-10 dataset. In terms of the number of epochs as well, the proposed density aware triplet loss requires much lesser number of epochs (98 epochs), whereas the triplet loss, triplet center loss, and the hard triplet loss takes 192, 144, and 149 epochs, respectively. 
\subsection{Convergence}

\begin{figure}[t]
     	\begin{center}
     		\includegraphics[width=.96\linewidth]{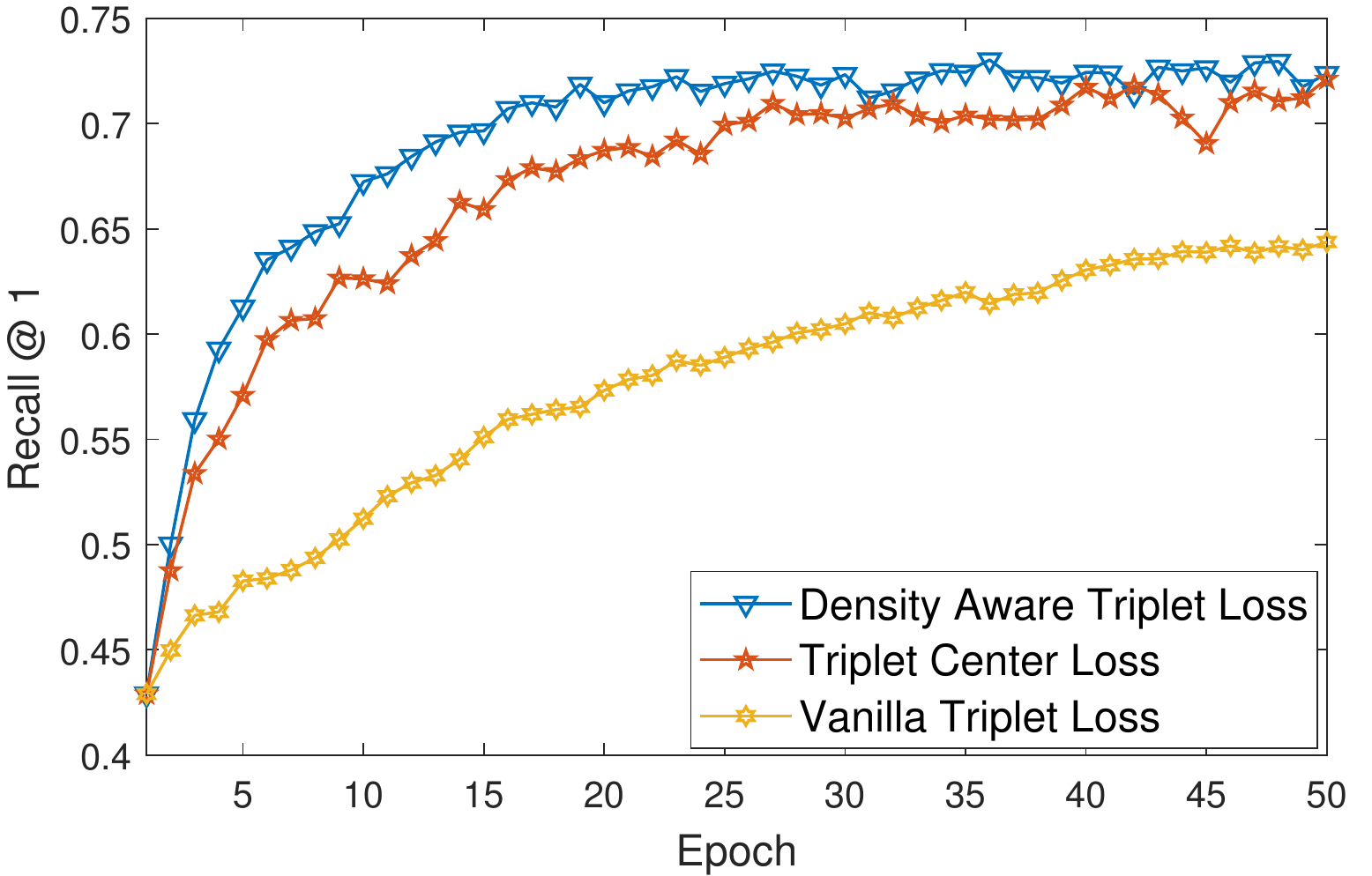}
     	\end{center}
      \vspace{-8pt}
     	\caption{Convergence of the proposed DATL compared with the vanilla triplet loss and triplet center loss on the STL-10 dataset.}  
     	
       \vspace{-8pt}
        \label{trip_conv}
     \end{figure}
The foremost advantage of the proposed density based deep metric learning approach is its ability to converge quickly as compared to the vanilla triplet and quadruplet loss methods. In addition, the proposed algorithm also converges much faster with respect to the triplet center loss~\cite{he2018triplet} which uses the centroid of the cluster in the loss function (Equation~\ref{modified_metric}). The proposed method avoids outliers and thus updates the weights of the model in such a way, so as to create embeddings in the most dense region of the clusters. This avoids large weight updates as the embeddings need not be shifted away from the dense region. As shown in Figure~\ref{trip_conv}, the convergence of the proposed density aware triplet loss (on a validation set which is prepared by randomly selecting 10\% samples from the training set) is superior than the vanilla triplet loss and the triplet center loss. For all the methods, convergence
is defined as the stage when the validation accuracy does not improve for 50 epochs at a stretch.

\begin{figure}[t]
     	\begin{center}
     		\includegraphics[width=.96\linewidth]{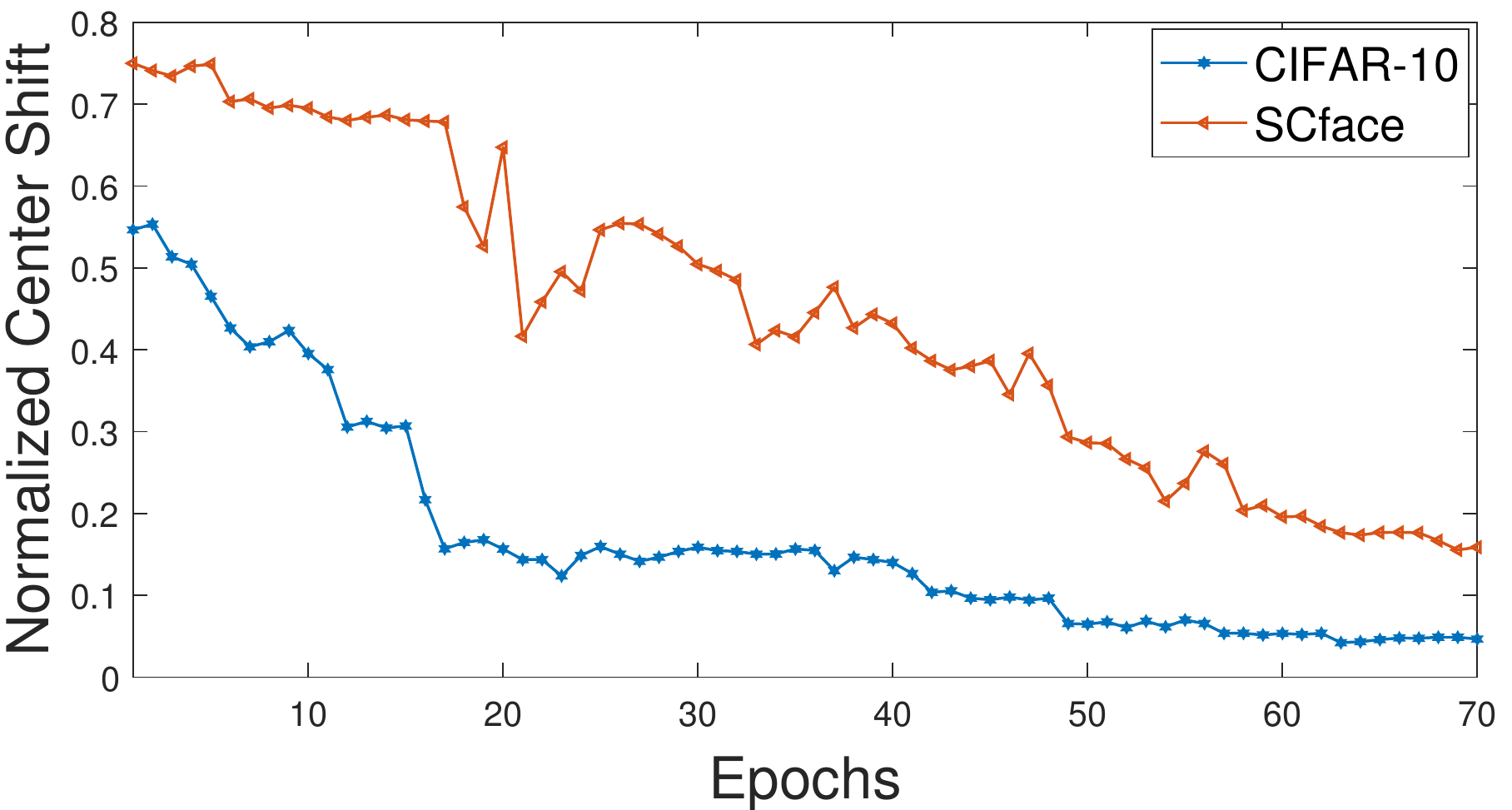}
     	\end{center}
      \vspace{-8pt}
     	\caption{Normalized center shift epoch-wise for the CIFAR-10 and SCface database (best viewed in color).}  
     	\vspace{-8pt}
    
        \label{center}
     \end{figure}
\subsection{Shifting of the Center}

The proposed approach iteratively evaluates the center towards the most dense region of each class (Figure~\ref{noisy_center}). An analysis is performed showing the magnitude of center shift for each epoch. The average of the center shift for all the classes is used to plot the graph in Figure~\ref{center} which shows that the magnitude of the center shift is much higher for the SCface database which has a very large number of noisy training samples. It can be also observed that the magnitude of the center shift decreases as the training progresses, thereby producing more compact clusters.   
      



\section{Conclusion}

The paper presents an elegant approach for density aware deep metric learning. The proposed approach can be augmented with any  deep metric learning technique such as triplet and quadruplet loss, and its variants. It results in superior convergence and accuracies, thus providing an important enhancement in current deep metric learning strategies. The proposed DATL and DAQL have also shown to be resilient to noisy training data compared to other deep metric learning methods. Extensive experiments on four datasets showcase the superiority of the proposed DATL and DAQL over existing deep metric learning techniques. 

\section{Acknowledgements}
M. Vatsa and R. Singh are partly supported by the Infosys Center for AI, IIIT Delhi. M. Vatsa is also supported through the Swarnajayanti Fellowship by Government of India. S. Ghosh is supported by TCS Research Fellowship.

{\small
\bibliographystyle{ieee_fullname}
\bibliography{egbib}
}
\end{document}